\documentclass{INTERSPEECH2023}
\interspeechcameraready

\title{Record Deduplication for Entity Distribution Modeling in ASR Transcripts}
\name{Tianyu Huang, Chung Hoon Hong, Carl Wivagg, Kanna Shimizu}
\address{Alexa AI, Amazon.com Inc, Boston, MA, USA}
\email{\{htianyu, honchung, cawivagg, kannashi\}@amazon.com}

\begin{document}

\maketitle
 
\begin{abstract}
Voice digital assistants must keep up with trending search queries. We rely on a speech recognition model using contextual biasing with a rapidly updated set of entities, instead of frequent model retraining, to keep up with trends. There are several challenges with this approach: (1) the entity set must be frequently reconstructed, (2) the entity set is of limited size due to latency and accuracy trade-offs, and (3) finding the true entity distribution for biasing is complicated by ASR misrecognition. We address these challenges and define an entity set by modeling customers’ true requested entity distribution from ASR output in production using record deduplication, a technique from the field of entity resolution. Record deduplication resolves or deduplicates coreferences, including misrecognitions, of the same latent entity. Our method successfully retrieves 95\% of misrecognized entities and when used for contextual biasing shows an estimated 5\% relative word error rate reduction.
\end{abstract}

\section{Introduction}
\label{introduction}

Voice digital assistants perform automatic speech recognition (ASR), natural language understanding (NLU), and entity resolution (ER) over a variety of domains, such as smart home, Q\&A, and entertainment. ASR serves a key role as it is upstream to all other functions, and its performance can determine how successfully the voice assistant meets customer needs. In modern systems, the ASR component is an ``end-to-end'' model based on architectures such as recurrent neural network transducers (RNN-T) \cite{JaitlyRnnt} or listen-attend-spell (LAS) \cite{VinaylsLas}. End-to-end ASR models map speech directly from voice data to graphemes without an intermediate phoneme step. As a consequence, relative to component-based models, end-to-end models struggle particularly with proper nouns and rare words \cite{ChiuLexicon}.

In this work, we focus on optimizing an end-to-end ASR model in the entertainment domain, where customers search for artists, songs, TV shows, and movies to play back. In entertainment, proper nouns and rare words are common with requests such as ``play Metro Boomin'' (a popular singer) or ``search for Bridgerton'' (a popular Netflix show). Therefore, one key challenge is spoken entity recognition, where we require speech recognition to perform at scale for these proper nouns, many of which can be easily confused with more common words. Furthermore, our model must rapidly adapt to a non-stationary distribution of requests where new hit songs, movies, and shows come out every week and may rocket to viral popularity levels with little warning. In this situation, we need to reconcile the time, cost, and engineering challenges of frequent model retraining and production release against the need to match the fast shifting customer request distribution.

We address these challenges with a twofold strategy. First, we leverage contextual biasing, where we alter the output probability of recognized entities without the latency or computational expense of full model retraining. This is accomplished by shallow fusion, which uses on-the-fly rescoring to adjust output probabilities during runtime inference with an externalized list of scored entities \cite{PangShallowFusion}. Updating biasing in this way does not require potentially costly model retraining and redeployment. However, contextual biasing is not without limitations: lengthy lists of entity names may degrade model accuracy or increase inference latency, which is critical for a fast response to a customer request. Consequently, biasing lists are limited to an entity budget of a few thousand entities. A few thousand does not even meet the number of new songs released every week, much less the entirety of other entertainment media content such as new movies, TV shows, and podcast episodes. 

The second part of our strategy is to optimize the utility of the entity budget and to best capture the non-stationary aspects of the request distribution. We accomplish this by starting with the observed entity distribution in the ASR output stream. The stream is a distorted version of the actual customer request distribution because of misrecognitions. For example, \textit{Archive 81}, a TV show, can been misrecognized as “arcade eighty one”, “r. kelly one”, or several other results depending on the speaker conditions.  We deduplicate these multiple references to the same entity by leveraging clustering techniques developed for record deduplication in ER research. After we correct for the distortions and reconstruct the entity distribution, we select entities and optimize weights to bias towards frequently misrecognized entities.

\section{Related Work}
\label{relatedwork}

We are aware of relatively little previous work to optimize the use of a shallow fusion entity budget. In early shallow fusion research, a weighted FST representing a language model is derived from training data or some other external reference \cite{PangShallowFusion, EvermannEarlyShallow}; however, the assumption that the entity distribution -- or entity language model -- can be derived from training data or other external data sources does not hold for practical applications. Thus, more recently, one group used a forecasting model to predict entity popularity and preemptively populate a biasing list \cite{OparinApple}. However, this approach limited itself to forecasting trends based on exact matches of the entity name to the reference in the user request. Thus, there is a particular risk of missing the entities most in need of contextual biasing: those for which the ASR output frequently differs from the entity name. In our work, we explicitly model the entity distribution, allowing us in principle to identify the most popular entities regardless of how they are referred to in user requests.

A subsequent line of research \cite{VanGyselDiscriminative} used comparison of entity references to entities in knowledge graphs to enhance ASR performance, but this differs in many respects from our approach of co-comparison of entity references to other references, and ultimately was not used to model the entity distribution for contextual biasing.

\begin{figure}
\centering
    \includegraphics[width=0.46\textwidth]{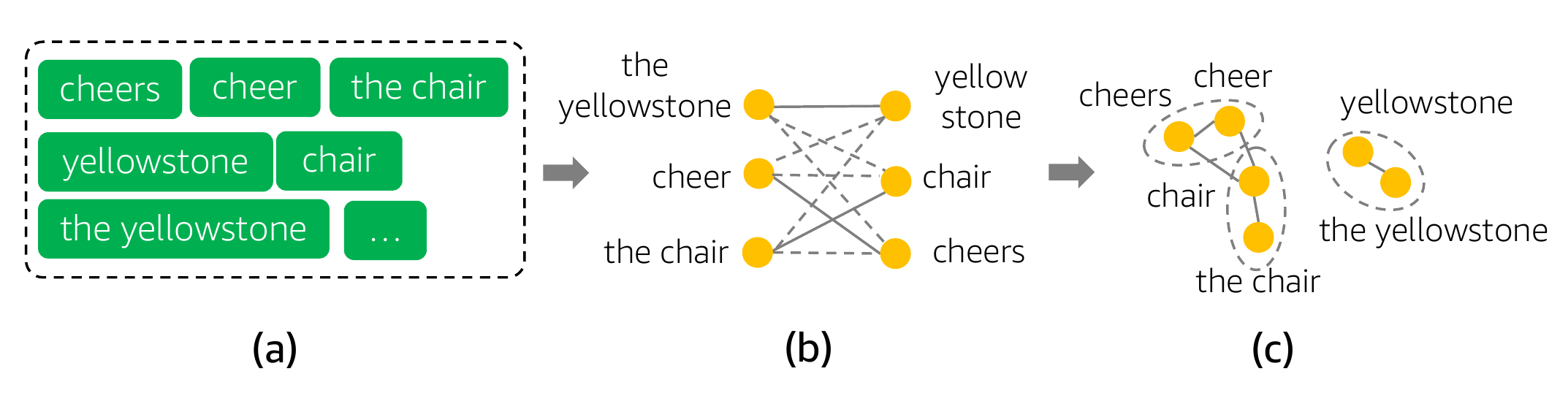}
    \caption{Record Deduplication Model Design. a) Blocking b) Entity Comparison c) Clustering.}
    \label{fig:approach1}
\end{figure}

\section{Model Design}
\label{sec:asdf}

Our model uses a record deduplication framework. We aim to identify different references to video title entities, like \textit{Bridgerton} or \textit{Archive 81}, in ASR outputs. Different ASR outputs may refer to the same entity because of ASR errors. For instance, ``arcade eighty one'' is likely a reference to \textit{Archive 81} just as ``archive eighty one'' is. So these request patterns are duplicates from the perspective of the entity being requested. In ER terms, the two request patterns are coreferent, a term we will use repeatedly going forward. We aim to deduplicate the total set of references to video title entities by clustering the references to each entity.

Formally, for a set $R$ of unique entity references drawn from a non-unique set of user requests $U$ each containing one reference, we seek to create subsets of $R$ corresponding to coreferences to some latent set of entities $E$. If the subset of references $r_1, r_2, ..., r_n$ are coreferent to entity $e_x\!\in\!E$ and $p_{r_i}$ is the probability of encountering $r_i$ in $U$, then $p_{e_x}=\sum_{i=1}^np_{r_i}$ and is the probability of a randomly selected user request referring to the entity $e_x$. The set of all $p_e$ is the latent entity distribution. 

\subsection{Acquisition of References from the ASR Model}
\label{sec:asr}

The record deduplication model takes as inputs the ASR outputs from our in-house RNN-T model, which consists of an encoder LSTM and prediction network layer with an embedding layer. The model also has a shallow fusion-based language model that is used to bias to global and personalized catalogs \cite{9413962}. The model was trained on over 200k hours of interactions with a voice assistant \cite{guo20_interspeech}. We use this model to get the $n$-best ASR recognitions from user requests or synthetic voice samples. In practice, ASR outputs are usually further processed by an NLU model to segment entity references and assign them to entity classes, but for this work, we confined our attention to a single entity class (video content titles) and to requests consisting only of references with no surrounding text.

\subsection{Record Deduplication Model}
\label{sec:recorddedup}

The record deduplication model transforms the ASR outputs, i.e., the $n$-best transcriptions of voice requests, into clusters of entity references, each of which is expected to contain references to exactly one entity. Record deduplication consists of three discrete steps: blocking, comparison, and clustering (Figure \ref{fig:approach1}). Blocking places references that are likely to be coreferences into a ``block'', or group of potential matches. The comparison task, which is analogous to other comparison, matching, and linking tasks in ER, consists of an all-against-all comparison of the set of potential coreferences in a block. The comparison model, which produces a $[0,1]$-bounded estimate of whether two references are linked, can be either a rule-based system or a machine learning model. Finally, clustering thresholds the similarity output, producing a final decision on which references are coreferent.

\subsubsection{Blocking}
\label{sssec:blocking}
We group our video title requests as a single block for input to comparison. However, in principle and without any methodological changes, our model could be extended to handle user requests for different types of content entities by repeating the procedure with a new block for each type identified in the request set, which could be done by an NLU model that performs entity recognition and classification.

\subsubsection{Comparison Model Features}
\label{sssec:modelfeatures}

For some block $B$ containing $n$ references $r_1, r_2, ..., r_n$, we next perform comparison: we construct a similarity matrix $S$ where element $s_{ij}$ is the similarity of $r_i$ and $r_j$, for some $r_i, r_j\!\in\!B$. In this work, we investigate several methods of computing $s_{ij}$. In all cases $s_{ij}$ is on the interval $[0,1]$.

All of our model configurations use ASR $n$-best cooccurrence. A list of ASR $n$-best candidate recognitions is automatically produced as part of the output of most ASR systems. The ASR $n$-best cooccurrence rate is a language-agnostic proxy for phonetic similarity. For a given $r_i$ and $r_j$, tabulate $c_{ij}$, the mean rate of cooccurrence per occurrence, using $c_{ij}=\big(p(r_i|r_j)+p(r_j|r_i)\big)/2$.

ASR $n$-bests have been used in a variety of contexts to improve model output; in modern ASR systems, they are understood to contain potentially relevant information for accurate recognition \cite{LinGrammticalErrorCorrection, RaghunathanAsrNbest}. Importantly, our system looks across aggregate cooccurrences and so is resilient to noise.

We also computed similarities between references from the perspective of cooccurrence in user histories (the item-item matrix in a collaborative filtering framework \cite{Linden2003Amazon}). For any pair of references $r_i$, $r_j$, their item similarity $u_{ij}$ is defined as $u_{ij} = (U_i \cdot U_j)/(\lvert\lvert U_i \rvert\rvert \cdot \lvert\lvert U_j \rvert\rvert)$, where $U_i$ is the frequency vector of requested references aggregated from all users who requested content by reference $r_i$ over a predefined time length, i.e., $U_i=(n_{1, i}, n_{2, i}, ...)$ where \(n_{j, i}=\sum_{k} v_k^j\) and $v_k^j$ is the number of times when the $k$-th user who requested $r_i$, also requested $r_j$. Intuitively, the score is measuring the similarity between the users who made the individual references, and the underlying assumption is that misrecognitions come from a relatively similar group of users to those with correct recognitions.

\subsubsection{Comparison Model Training}
\label{sssec:modeltraining}
Our initial proof of concept model uses ASR $n$-best cooccurrence alone to measure similarity. For more advanced models, to weigh ASR $n$-best cooccurrence and item similarity, we formulated the comparison task as a binary classification problem and trained a machine learning model to provide similarity scores on the $[0, 1]$ continuum for candidate pairs of entity references.

The training dataset consists of ASR and item similarities derived from pairs of references labeled $1$ if they are coreference and $0$ otherwise. We mine the ground truth labels from user interactions with our digital assistant. When a user selects an item from a list of displayed results after a query, we observe whether the most-clicked item has a clickthrough rate (average number of clicks per impression) greater than 50\%. If so, we consider the query unambiguous in the sense that it is intended to target a specific video title (e.g., ``star wars a new hope'') as opposed to a broad search (e.g., ``star wars movies''), and label the query and entity name as a reference pair. The positive training data are made up of (1) query pairs with the same intended video and (2) the same query as both members of the pair. The negative training data are created by sampling from query pairs with different intended videos. We use an equal amount of positive and negative samples for model training.

We used several classification algorithms to train the comparison models, including logistic regression, decision tree, and support vector classification \cite{alpaydin2020introduction} \footnote{The models are trained using the \textit{scikit-learn} package \cite{DuchesnayScikit} and the training parameters are kept to module defaults.}. All models are trained with an 80\%/20\% train-test data split. Note that the goal of the comparison model is to give binary decisions connecting similar query pairs, analogous to the link prediction task \cite{Zhang2018Link} in network theory for cluster detection on the generated graph \cite{Barbieri2014Who}.

\subsubsection{Clustering Algorithm}
\label{sssec:clusteringmodel}

For the proof-of-concept model using similarity from ASR $n$-bests, we convert the similarity matrix $S$ to an adjacency matrix $A$ with $a_{ij} \in \{0, 1\}$ via thresholding, with the threshold empirically tuned based on the training data. For the classifier models, $A$ is constructed directly from the classifier output labels, which are also in $\{0, 1\}$. The sets of adjacent elements form clusters $c_1, c_2, ..., c_m$ with $m \leq \lvert R \rvert$. For some cluster $c_i$, if $r_x \in c_i$ and $r_y \in c_i$, we conclude that $r_x$ and $r_y$ are coreferent.

\subsection{Character Edit Baseline}
\label{ssec:modelbaseline}

To contrast record deduplication with a simpler approach, we performed similarity comparison between ASR inputs and outputs from the public dataset using character edit distance. If the distance between an ASR output and its respective input was less than or equal to the next closest edit distance to other ASR inputs, we considered these entities matched (true positive). Otherwise, we considered them mismatched and counted both a false positive for the entity the ASR output was matched to and a false negative for the entity it was not associated with.

\subsection{Application to Shallow Fusion}
\label{ssec:shallowfusionapplication}

Simply boosting the top $k$ values in the set of all $p_e$ would provide an improvement over an entity distribution derived from static training data. In practice, we use user feedback signals to determine which references to each latent entity are likely misrecognitions and perform contextual biasing on the entity names most likely to be both requested and misrecognized.

\section{Data and Evaluation}
\label{sec:data}

We use two input datasets: a publicly reproducible one using Amazon Polly and an in-house one derived from anonymized/de-identified real user interactions with our digital assistant in English. The in-house data was required because the item-item matrix will not have meaningful information if computed from synthetic interactions. We thus use the public dataset for an initial proof-of-concept model using only the ASR $n$-best cooccurrence feature described in Section \ref{sssec:modelfeatures} and then move to the anonymized/de-identified in-house dataset for more complicated models involving item-item cooccurrence.

Since the publicly reproducible dataset is generated from known text, the ground truth to compute model accuracy is readily available. For the anonymized/de-identified in-house dataset, we rely on sets of coreferent pairs identified by user behavior, as described in Sections \ref{ssec:recalldata} and \ref{ssec:precisiondata}.

\subsection{Public Dataset}
\label{ssec:publicdata}


We generated 900 synthetic voice samples \cite{Huang_Open_dataset_used_2023} using Amazon Polly \cite{amazon_polly} on randomly selected movie titles from the MovieLens 25M Dataset \cite{grouplens_2021}. Before inputting the data for audio synthesis, we preprocessed and added tokenization operations to the movie titles to change official movie titles into spoken forms (e.g., ``Tiny Times III'' to ``tiny times three'') to reflect speech patterns of voice digital assistant consumers. We used nine different voice profiles from English (US) language variants available on Amazon Polly to generate nine samples for each entity. For the audio synthesis, neural text-to-speech engine was chosen to create high-quality audio streams. Amazon Polly's synthesize-speech command was used to generate the audio and convert it to a wav file.

\subsection{Recall Computation}
\label{ssec:recalldata}

Since we do not have ground truth for our anonymized/de-identified in-house dataset, we instead rely on feedback extracted from user interaction sessions. To identify related ASR output variants that should be clustered together (or deduplicated) in record deduplication, we consider cases where a user request for some reference $r_a$ to entity $e_x$ does not result in a satisfactory response, and as a consequence, the user repeats the request, perhaps with clearer enunciation or in a louder voice, resulting in ASR recognition variant reference $r_b$. If many users repeat a given pair $(r_a, r_b)$, we can conclude that the two variants are coreferent to $e_x$. Counting the number of such pairings recalled as edges in the record deduplication cluster output gives us an estimate of the model's sensitivity: these pairings are true positives, while known edges that could have been output but were not are false negatives.

The recall metrics do not equate to word or sentence error rates for ASR models, since they describe the \emph{relative} improvement in ASR sentence error rate.

\subsection{Precision Computation}
\label{ssec:precisiondata}

We identify false positive reference pairs output by the record deduplication model by treating reference-item/entity pairs in Section \ref{sssec:modeltraining} as ground truth: to calculate precision, if $r_a$ and $r_b$ are clustered together in record deduplication outputs, but we have $r_a$ resolved to $e_x$ with high user satisfaction and $r_b$ resolved to $e_y, y \neq x$, then $(r_a, r_b)$ is counted as a false positive. The precision is the total number of edges output minus such false positives divided by the total number of edges output for which both entities have ER results with positive user feedback.

Although it is possible that a reference could resolve to two entities, this is rare in practice ($<0.1\%$).

\section{Results and Discussion}
\label{sec:results}

We performed two sets of experiments: an initial proof-of-concept using an open dataset and similarity as measured by ASR $n$-best cooccurrence, and then a more extensive set of comparisons of different possible comparison models containing both of the ASR $n$-best feature and the item similarity feature described in Section \ref{sssec:modelfeatures}.

\subsection{Initial Proof of Concept}
\label{ssec:nbestresults}

ASR $n$-best cooccurrence by itself achieved a recall of $0.997$ on the synthetic dataset of ASR errors created using Amazon Polly (Table \ref{pocresultstable}), with no loss of precision. In other words, it successfully grouped misrecognized references with correctly recognized coreferents $99.7\%$ of the time, without incorrectly grouping any references to different entities. By comparison, simply using a nearest neighbors approach to grouping misrecognized ASR variants with correctly recognized coreferences recalled only $50.0\%$ of errors, with a significant cost to precision. However, these results translated poorly to the anonymized/deidentified in-house dataset based on live traffic for our digital assistant. For this dataset, the method achieved a recall of only $0.918$, with a precision of $0.913$.

The loss in performance likely arises from several factors. First, in the more limited public dataset, we were able to use an ASR $n$-best with $n=5$, while the cost and latency requirements on the large anonymized/deidentified in-house dataset, which was generated at runtime, required $n\leq2$.

Additional performance loss likely results from the substantially wider distribution of errors encountered in live traffic. Live recordings may also contain misdirected traffic from the NLU model performing the blocking.

\begin{table}[!h]
    \caption{\label{pocresultstable}ASR $n$-Best Model Record Deduplication Results}
    \begin{center}
    \begin{tabular}{ lllll }
     \toprule
     \cmidrule{1-5}
     Dataset & Model & Recall & Precision & $F_1$ \\ 
    \midrule
     Public & RNN-T only & 0 & 1 & 0.000 \\
     Public & Edit similarity & 0.500 & 0.455 & 0.476 \\
     Public & Record dedup. & 0.997 & 1.000 & 0.998 \\
     Anon. Real & Record dedup. & 0.922 & 0.913 & 0.917 \\ 
    \bottomrule
    \end{tabular}
    \end{center}
\end{table}

\subsection{Results Including Item-Item Cooccurrence}

To improve accuracy, we introduced the item-item similarity feature. The hypothesis behind this feature is that users making malformed or misrecognized entity references should be similar to users requesting the same entity through a correctly recognized reference. In other words, we hypothesized that distinct groups of users will request particular entities, and that request misrecognition is to some extent random within each group for a particular entity.

We produced comparison scores using a linear weighting of the two features as well as tree- and SVM-based models (Table \ref{mainresultstable}). The linear and tree-based models recovered 36\% and 49\% respectively of the $F_1$ loss in moving to live data, but the SVM only recovered 20\% of performance, probably because of its higher dependence on hyperparameter tuning.

\begin{table}
    \caption{\label{mainresultstable} Combined $n$-Best/Item-Item Cooccurrence Model Record Deduplication Results on Anonymized In-House Dataset}
    \begin{center}
    \begin{tabular}{ llll }
     \toprule
     \cmidrule{1-4}
     Model & Recall & Precision & $F_1$ \\ 
     \midrule
     $n$-best-only & 0.922 & 0.913 & 0.917 \\ 
     Linear & 0.934 & 0.958 & 0.946 \\ 
     Tree & 0.954 & 0.959 & 0.957 \\
     SVM & 0.970 & 0.899 & 0.933 \\
     \bottomrule
    \end{tabular}
    \end{center}
\end{table}

\subsection{Results on Live Data}

The cluster output provides a model of entity requests; combined with system logs, we can quantify the number of entity requests in each cluster that led to the user being served the correct entity (because the ASR output was sufficiently similar to the canonical form that downstream ER could perform correctly). We utilize the record deduplication's model of traffic to perform contextual biasing on the runtime ASR model (Table \ref{liveresultstable}), boosting the effective LM probability of outputting canonical entities as in \cite{9413962} over incorrectly resolved variants. For evaluation, we use machine generated transcripts from record deduplication as reference transcripts. Performing contextual boosting on entities from record deduplication shows relative word error rate (WER) reduction of 0.67\% over randomly selected, anonymized, and annotated user utterances. By comparison, using a selection of the top $k$ most-mentioned entities actually \emph{increased} the error rate by 2.78\%. This potentially surprising finding is consistent with the increase in WER seen for the ``popular last week'' heuristic in \cite{VanGyselDiscriminative}. The modest size of the record deduplication result likely arises from the high diversity of entities in our total dataset. In a smaller selection (``modeled only'' in Table \ref{liveresultstable}) of utterances that each contain a coreference used in record deduplication, the relative improvement was 13.01\%. This figure is comparable to the WER reductions reported in \cite{9413962} using a different approach, but the data distributions are very different. Extrapolating from the 0.67\% relative WER by a factor of $1/0.1301$, the entities selected for boosting account for roughly 5\% of the misrecognitions in our live distribution.


\begin{table}
    \caption{\label{liveresultstable} Relative WER Results on Anonymized In-House Dataset}
    \begin{center}
    \begin{tabular}{ lll }
     \toprule
     \cmidrule{1-3}
     Model & Refs in Dataset & rel. WER (\%) \\ 
     \midrule
     base & full & 0 \\
     base + TopK entities & full & 2.78 \\
     base + Record dedup. & full & -0.67 \\
     base & modeled only & 0 \\
     base + Record dedup. & modeled only & -13.01 \\ 
     \bottomrule
    \end{tabular}
    \end{center}
\end{table}

\section{Conclusion}
\label{sec:conclusion}

Our work demonstrates that using a comparison model within an entity deduplication framework is an effective way of building a model of entity requests. Using this entity distribution and the information it contains about frequently misrecognized entities, we can utilize a shallow fusion entity budget more effectively than a naive baseline.

Two promising ways to develop the record deduplication model are (1) the use of a deep phonetic similarity model, similar to \cite{CampbellSiam} to improve performance in the comparison task and (2) using community detection approaches like spectral clustering \cite{FORTUNATO201075} for the clustering task.



\bibliographystyle{IEEEtran}
\bibliography{template}

\begin{thebibliography}{10}
\providecommand{\url}[1]{#1}
\csname url@samestyle\endcsname
\providecommand{\newblock}{\relax}
\providecommand{\bibinfo}[2]{#2}
\providecommand{\BIBentrySTDinterwordspacing}{\spaceskip=0pt\relax}
\providecommand{\BIBentryALTinterwordstretchfactor}{4}
\providecommand{\BIBentryALTinterwordspacing}{\spaceskip=\fontdimen2\font plus
\BIBentryALTinterwordstretchfactor\fontdimen3\font minus
  \fontdimen4\font\relax}
\providecommand{\BIBforeignlanguage}[2]{{%
\expandafter\ifx\csname l@#1\endcsname\relax
\typeout{** WARNING: IEEEtran.bst: No hyphenation pattern has been}%
\typeout{** loaded for the language `#1'. Using the pattern for}%
\typeout{** the default language instead.}%
\else
\language=\csname l@#1\endcsname
\fi
#2}}
\providecommand{\BIBdecl}{\relax}
\BIBdecl

\bibitem{JaitlyRnnt}
A.~Graves and N.~Jaitly, ``Towards end-to-end speech recognition with recurrent
  neural networks,'' in \emph{Proceedings of the 31st International Conference
  on International Conference on Machine Learning - Volume 32}.\hskip 1em plus
  0.5em minus 0.4em\relax JMLR.org, 2014, p. II–1764–II–1772.

\bibitem{VinaylsLas}
W.~Chan, N.~Jaitly, Q.~Le, and O.~Vinyals, ``Listen, attend and spell: A neural
  network for large vocabulary conversational speech recognition,'' in
  \emph{2016 IEEE International Conference on Acoustics, Speech and Signal
  Processing (ICASSP)}, 2016, pp. 4960--4964.

\bibitem{ChiuLexicon}
\BIBentryALTinterwordspacing
T.~N. Sainath, R.~Prabhavalkar, S.~Kumar, S.~Lee, A.~Kannan, D.~Rybach,
  V.~Schogol, P.~Nguyen, B.~Li, Y.~Wu, Z.~Chen, and C.-C. Chiu, ``{No Need for
  a Lexicon? Evaluating the Value of the Pronunciation Lexica in End-to-End
  Models},'' in \emph{2018 IEEE International Conference on Acoustics, Speech
  and Signal Processing (ICASSP)}.\hskip 1em plus 0.5em minus 0.4em\relax IEEE
  Press, 2018, p. 5859–5863. [Online]. Available:
  \url{https://doi.org/10.1109/ICASSP.2018.8462380}
\BIBentrySTDinterwordspacing

\bibitem{PangShallowFusion}
D.~Zhao, T.~N. Sainath, D.~Rybach, P.~Rondon, D.~Bhatia, B.~Li, and R.~Pang,
  ``{Shallow-Fusion End-to-End Contextual Biasing},'' in \emph{Proc.
  Interspeech 2019}, 2019, pp. 1418--1422.

\bibitem{EvermannEarlyShallow}
R.~Huang, O.~Abdel-hamid, X.~Li, and G.~Evermann, ``{Class LM and Word Mapping
  for Contextual Biasing in End-to-End ASR},'' in \emph{Proc. Interspeech
  2020}, 2020, pp. 4348--4351.

\bibitem{OparinApple}
\BIBentryALTinterwordspacing
C.~Van~Gysel, M.~Tsagkias, E.~Pusateri, and I.~Oparin, ``Predicting entity
  popularity to improve spoken entity recognition by virtual assistants,'' in
  \emph{Proceedings of the 43rd International ACM SIGIR Conference on Research
  and Development in Information Retrieval}.\hskip 1em plus 0.5em minus
  0.4em\relax New York, NY, USA: Association for Computing Machinery, 2020, p.
  1613–1616. [Online]. Available:
  \url{https://doi.org/10.1145/3397271.3401298}
\BIBentrySTDinterwordspacing

\bibitem{VanGyselDiscriminative}
M.~Saebi, E.~Pusateri, A.~Meghawat, and C.~V. Gysel, ``{A Discriminative
  Entity-Aware Language Model for Virtual Assistants},'' in \emph{Proc.
  Interspeech 2021}, 2021, pp. 2032--2036.

\bibitem{9413962}
A.~Gourav, L.~Liu, A.~Gandhe, Y.~Gu, G.~Lan, X.~Huang, S.~Kalmane, G.~Tiwari,
  D.~Filimonov, A.~Rastrow, A.~Stolcke, and I.~Bulyko, ``Personalization
  strategies for end-to-end speech recognition systems,'' in \emph{ICASSP 2021
  - 2021 IEEE International Conference on Acoustics, Speech and Signal
  Processing (ICASSP)}, 2021, pp. 7348--7352.

\bibitem{guo20_interspeech}
J.~Guo, G.~Tiwari, J.~Droppo, M.~V. Segbroeck, C.-W. Huang, A.~Stolcke, and
  R.~Maas, ``{Efficient Minimum Word Error Rate Training of RNN-Transducer for
  End-to-End Speech Recognition},'' in \emph{Proc. Interspeech 2020}, 2020, pp.
  2807--2811.

\bibitem{LinGrammticalErrorCorrection}
L.~Zhu, W.~Liu, L.~Liu, and E.~Lin, ``{Improving ASR Error Correction Using
  N-Best Hypotheses},'' in \emph{2021 IEEE Automatic Speech Recognition and
  Understanding Workshop (ASRU)}, 2021, pp. 83--89.

\bibitem{RaghunathanAsrNbest}
A.~Raghuvanshi, V.~Ramakrishnan, V.~Embar, L.~Carroll, and K.~Raghunathan,
  ``{Entity resolution for noisy ASR transcripts},'' in \emph{Conference on
  Empirical Methods in Natural Language Processing}, 2019.

\bibitem{Linden2003Amazon}
G.~Linden, B.~Smith, and J.~York, ``Amazon.com recommendations: item-to-item
  collaborative filtering,'' \emph{IEEE Internet Computing}, vol.~7, no.~1, pp.
  76--80, 2003.

\bibitem{alpaydin2020introduction}
E.~Alpaydin, \emph{Introduction to machine learning}.\hskip 1em plus 0.5em
  minus 0.4em\relax MIT press, 2020.

\bibitem{DuchesnayScikit}
F.~Pedregosa, G.~Varoquaux, A.~Gramfort, V.~Michel, B.~Thirion, O.~Grisel,
  M.~Blondel, P.~Prettenhofer, R.~Weiss, V.~Dubourg, J.~Vanderplas, A.~Passos,
  D.~Cournapeau, M.~Brucher, M.~Perrot, and E.~Duchesnay, ``Scikit-learn:
  Machine learning in {P}ython,'' \emph{Journal of Machine Learning Research},
  vol.~12, pp. 2825--2830, 2011.

\bibitem{Zhang2018Link}
M.~Zhang and Y.~Chen, ``Link prediction based on graph neural networks,'' in
  \emph{Proceedings of the 32nd International Conference on Neural Information
  Processing Systems}.\hskip 1em plus 0.5em minus 0.4em\relax Red Hook, NY,
  USA: Curran Associates Inc., 2018, p. 5171–5181.

\bibitem{Barbieri2014Who}
\BIBentryALTinterwordspacing
N.~Barbieri, F.~Bonchi, and G.~Manco, ``Who to follow and why: Link prediction
  with explanations,'' in \emph{Proceedings of the 20th ACM SIGKDD
  International Conference on Knowledge Discovery and Data Mining}.\hskip 1em
  plus 0.5em minus 0.4em\relax New York, NY, USA: Association for Computing
  Machinery, 2014, p. 1266–1275. [Online]. Available:
  \url{https://doi.org/10.1145/2623330.2623733}
\BIBentrySTDinterwordspacing

\bibitem{Huang_Open_dataset_used_2023}
\BIBentryALTinterwordspacing
T.~Huang, H.~Chung~Hoon, C.~Wivagg, and K.~Shimizu, ``{Open dataset used in
  Record Deduplication for Entity Distribution Modeling in ASR Transcripts},''
  May 2023. [Online]. Available:
  \url{https://github.com/recorddeduplication/recorddeduplication}
\BIBentrySTDinterwordspacing

\bibitem{amazon_polly}
\BIBentryALTinterwordspacing
``{Amazon Polly},'' accessed: 2023-01-31. [Online]. Available:
  \url{https://aws.amazon.com/polly/}
\BIBentrySTDinterwordspacing

\bibitem{grouplens_2021}
\BIBentryALTinterwordspacing
``Movielens 25m dataset,'' Mar 2021. [Online]. Available:
  \url{https://grouplens.org/datasets/movielens/25m/}
\BIBentrySTDinterwordspacing

\bibitem{CampbellSiam}
\BIBentryALTinterwordspacing
X.~Zhou, R.~Bao, and W.~M. Campbell, ``Phonetic embedding for asr robustness in
  entity resolution,'' in \emph{Interspeech 2022}, 2022. [Online]. Available:
  \url{https://www.amazon.science/publications/phonetic-embedding-for-asr-robustness-in-entity-resolution}
\BIBentrySTDinterwordspacing

\bibitem{FORTUNATO201075}
\BIBentryALTinterwordspacing
S.~Fortunato, ``Community detection in graphs,'' \emph{Physics Reports}, vol.
  486, no.~3, pp. 75--174, 2010. [Online]. Available:
  \url{https://www.sciencedirect.com/science/article/pii/S0370157309002841}
\BIBentrySTDinterwordspacing

\end{thebibliography}

\end{document}